\pdfoutput=1

\documentclass{article}
\usepackage{spconf,amsmath,amssymb,float,multirow,graphicx,comment}
\usepackage{subcaption}
\usepackage{hyperref}
\usepackage{graphicx}
\usepackage{epsfig}

\DeclareMathOperator{\sign}{sign}
\DeclareMathOperator*{\argmax}{arg\,max}

\begin{document}\sloppy

\def\R{{\mathbf R}} 
\def\L{{\cal L}} 

\title{AdvHat: Real-world adversarial attack on ArcFace Face ID system}
\name{Stepan Komkov$^{1,2}$, Aleksandr Petiushko$^{1,2}$}
\address{
	$^1$ Lomonosov Moscow State University, MaTIS Chair\\
	$^2$ Huawei Moscow Research Center, Intelligent Systems Lab\\
	komkov.stepan@huawei.com, petyushko.alexander1@huawei.com
}

\maketitle

\begin{abstract}
In this paper we propose a novel easily reproducible technique to attack the best public Face ID system ArcFace in different shooting conditions. To create an attack, we print the rectangular paper sticker on a common color printer and put it on the hat. The adversarial sticker is prepared with a novel algorithm for off-plane transformations of the image which imitates sticker location on the hat. Such an approach confuses the state-of-the-art public Face ID model LResNet100E-IR, ArcFace@ms1m-refine-v2 and is transferable to other Face ID models.
\end{abstract}
\section{Introduction}
\label{sec:intro}

Last years face recognition systems based on deep learning and massive training data provided a very high level of recognition which outperforms human level of verification \cite{Face-Ver-Taigman} as well as identification \cite{Face-ID-Taigman}.

In the beginning, only the big corporations could afford training of the Face ID models on a huge amount of private data (e.g. 200M of faces in \cite{FaceNet}). Later, with the introduction of quite big public datasets (mostly CASIA-WebFace \cite{CASIA} and MS-Celeb-1M \cite{MS-Celeb-1M}) and new types of losses for training Face ID models (in particular, angular-based losses: L-Softmax \cite{FaceID-L-Softmax}, A-Softmax \cite{FaceID-A-Softmax}, AM-Softmax \cite{FaceID-AM-Softmax, FaceID-CosFace} and ArcFace \cite{ArcFace-paper}), even the models trained with public datasets by independent researchers can be of the same (or similar) performance as the proprietary models provided by large companies. For example, the ArcFace (Insightface) solution is comparable with Microsoft and Google models in such challenges as Megaface \cite{Megaface1, Megaface2}, NIST FRVT \cite{NIST-FRVT-1N}, and Trillion Pairs \cite{Deepglint}.

\begin{figure}[t]
	\begin{minipage}[t]{1.\linewidth}
		\center{\includegraphics[width=\columnwidth]{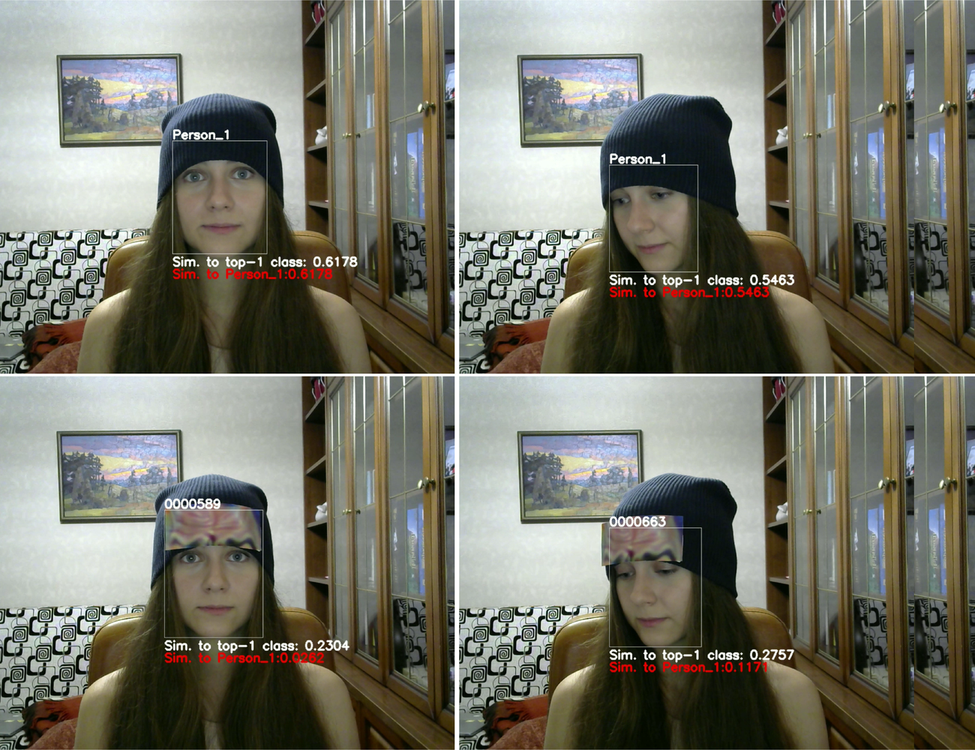}}
	\end{minipage}
	\caption{A novel approach to attack the Facial Recognition system. A sticker placed on the hat significantly reduces the similarity to the ground truth class. Similarity to the ground truth decreases by 0.592 on the left pair and by 0.429 on the right pair.}
	\label{fig:example}
\end{figure}

Nowadays, an increasing emphasis has been placed on the adversarial attacks on deep neural networks. One of the reasons for this is that adversarial attacks can be implemented in the real world \cite{AdvA-RealClass-Paper-Kurakin}. Recently, a form of adversarial attack \cite{AdvA-first} on previous generation Face ID models has been proposed \cite{AdvA-RealFaceID-Sharif1, AdvA-RealFaceID-Sharif2}. A drawback of the proposed methods is that you need to cut out complex shape object from the paper. Another drawback is that shooting conditions (lighting, angle of the face and background) were fixed. In this work we propose an easily reproducible (rectangular image, printed and stuck to a hat) practical adversarial attack called AdvHat on the best public Face ID model LResNet100E-IR, ArcFace@ms1m-refine-v2 \cite{ArcFace-model}. Demonstration of AdvHat is depicted in \textbf{Figure \ref{fig:example}}. The advantages of the proposed AdvHat can be summarized as follows:
\begin{itemize}
	\item We implemented a real-world adversarial attack on the state-of-the-art public Face ID system using sticker on the hat.
	\item The attack is easily reproducible. It is necessary to print only a color rectangle.
	\item One attack works in different shooting conditions. 
	\item We present a novel technique of the sticker projection to the image during the attack to make it real-like.
	\item The attack is transferable to other Face ID models.
\end{itemize}

The source code\footnote{\url{https://github.com/papermsucode/advhat}} and video demonstration\footnote{\url{https://youtu.be/a4iNg0wWBsQ}} are available on the Internet.

\section{Background And Related Works}
\label{sec:bkg}

Firstly we review the adversarial attacks as our work uses the concept of adversarial attack. Second, we touch the emerging area of adversarial attacks in the real world, or in other words practical attacks, since our work aims to construct an adversarial attack working under conditions of the real world (as a contrast to simple attack on pixels in the digital domain). 

\subsection{Adversarial attacks}

The whole concept of adversarial attacks is quite simple: let us slightly change the input to a classifying neural net so that the recognized class will change from correct to some other class (first adversarial attacks were made only on classifiers). The pioneering work \cite{AdvA-first} formulates the task as follows:

\begin{itemize}
	\item Minimize $||r||_{2}$ so as:
	\begin{enumerate}
		\item $f(x) = c_{gt}$,
		\item $f(x+r) = c_{t} \neq c_{gt}$,
		\item $x+r\in [0,1]^{m}$,
	\end{enumerate}
\end{itemize}
where $x\in [0,1]^{m}$ is an input to a classifier $f$, $c_{gt}$ --- correct ground truth class for $x$, $c_{t}\neq c_{gt}$ --- target class for $x+r$, $r\in[0,1]^{m}$ --- small perturbation to $x$ that we need to construct.   

Note that if we need to get the incorrect class, the attack is called untargeted (or dodging in face recognition cases), and if we need to get the specific predefined class $c_{t}$, the attack is called targeted (or impersonation in face recognition cases).

In \cite{AdvA-first} the authors propose to use a quasi-newton L-BFGS-B method to solve the task formulated above. Simpler and more efficient method called Fast Gradient-Sign Method (FGSM) is proposed in \cite{AdvA-FGSM}. This method suggests using the gradients with respect to the input and constructing an adversarial image using the following formula: $x = x + \epsilon\,\sign{\nabla_{x}\L(\theta, x, c_{gt})}$ (or $x = x - \epsilon\,\sign{\nabla_{x}\L(\theta, x, c_{t})}$ in case of targeted attack). Here $\L(\theta, x, y)$ is a loss function (e.g. cross-entropy) which depends on the weights of the model $\theta$, input $x$, and label $y$. Note that usually one step is not enough and we need to do a number of iterations described above each time using the projection to the initial input space (e.g. $x\in [0,1]^{m}$). It is called projected gradient descent (PGD) \cite{AdvA-PGD}.

It turns out that using momentum for the iterative procedure of an adversarial example construction is a good way to increase the robustness of the adversarial attack \cite{AdvA-Momentum}. 

All the aforementioned adversarial attacks suggest that we restrict the maximum per-pixel perturbation (in case of image as an input) i.e. use $L_{\infty}$ norm. Another interesting case is when we do not concentrate on the maximum perturbation but we strive to achieve the fewest possible number of pixels to be attacked ($L_{0}$ norm). One of the first examples of such attack is the Jacobian-based Saliency Map Attack (JSMA) \cite{AdvA-JSMA}, where the saliency maps are constructed of the pixels that are the most prone to cause the misclassification.

Another extreme case of attack for the $L_{0}$ norm is a one-pixel attack \cite{AdvA-OnePixel}. The authors use differential evolution for this specific case, the algorithm which lies in the class of evolutionary algorithms. 
 
It should be mentioned that not only classification neural nets are prone to adversarial attacks. There are also attacks for detection and segmentation \cite{AdvA-SegmDet}.

Another interesting property of the adversarial attacks is that they are transferable between different neural networks \cite{AdvA-first}. An attack prepared using one model can successfully confuse another model with different architecture and training dataset.

Usually, the adversarial attacks which are constructed using the specific architecture and even the weights of the attacked model are called white-box attacks. If the attack has no access to model weights then it is called a black-box attack \cite{AdvA-BB}.

Usually, attacks are constructed for the specific input (e.g. photo of some object). This is called an input-aware attack. Adversarial attacks are called universal when one successful adversarial perturbation can be applied for any image \cite{AdvA-Universal}.

In this work we concentrate on a white-box input-aware adversarial attack as the first step of our research.

\begin{figure*}[t]
	\begin{minipage}[h]{1.\linewidth}
		\center{\includegraphics[width=0.8\textwidth]{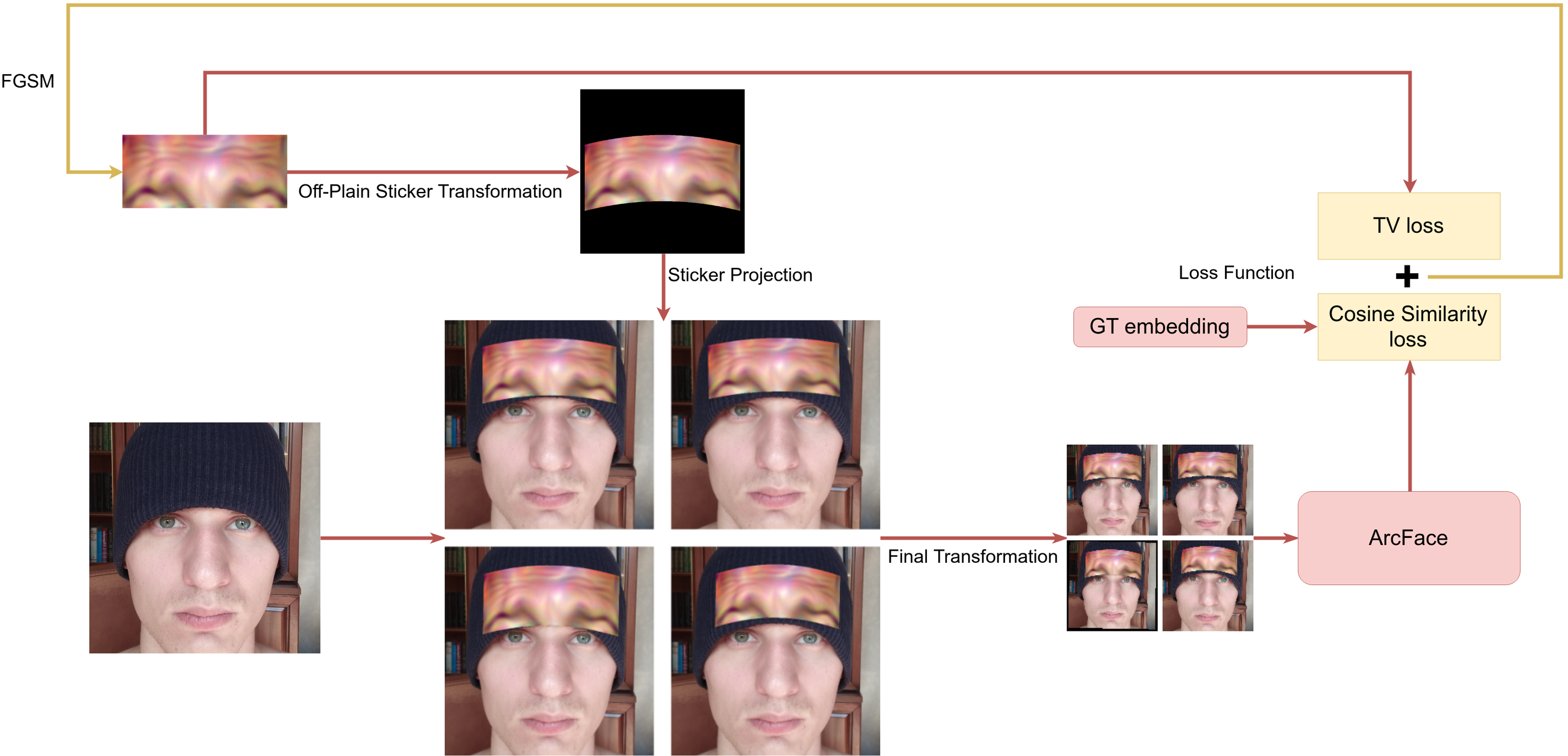}}
	\end{minipage}
	\caption{Schema of the whole pipeline of the attack. First, we reshape sticker to a real-look form. Second, we project it on the face images. Third, we transform images into the ArcFace input templates using slightly different parameters for the transformation. Finally, we feed templates to the ArcFace, evaluate cosine similarities and TV loss. Thus we can get the gradients signs which are used to modify the sticker image.}
	\label{fig:pipeline}
\end{figure*}

\subsection{Attacks in physical world}

Although adversarial attacks are quite successful in the digital domain (where we can change the image on the pixel level before feeding it to a classifier), in the physical (i.e. real) world the efficiency of adversarial attacks is still questionable. Kurakin \textit{et al.} demonstrate the potential for further research in this domain \cite{AdvA-RealClass-Paper-Kurakin}. They discovered that if an adversarial image is printed on the paper and then shot by a camera phone it still can successfully fool classification network.

It turns out that the most successful paradigm to construct the real-world adversarial examples is an Expectation Over Transformation (EOT) algorithm \cite{AdvA-RealClass-Transformations-Athalye}. This approach takes into account that in the real world the object usually undergoes a set of transformations (scaling, jittering, brightness and contrast changes, etc). The task is to find an adversarial example which is robust under this set of transformations $T$ and can be formulated as follows:

\begin{itemize}
	\item Find such $\argmax_{r}\mathbb{E}_{g\sim T}P(c_{t}|g(x+r))$ so as:
	\begin{enumerate}
		\item $f(x) = c_{gt} \neq c_{t}$,
		\item $\mathbb{E}_{g\sim T}||g(x+r) - g(x)||_{p} < \epsilon$,
		\item $x+r\in [0,1]^{m}$,
	\end{enumerate}
\end{itemize}
where we use the notion of $\epsilon$-vicinity in some $L_{p}$ space.

One of the first papers adopting this idea is Adversarial Patch \cite{AdvA-AdvPatch}. In this work the authors use EOT for a set of transformations including rotations and translations to construct the universal patch for the ImageNet \cite{ImageNet} classifier. It is noteworthy that the authors do not concentrate on $L_{p}$ norms for perturbations where $p\geq 1$ but on $L_{0}$ norm.

Another work with the usage of $L_{0}$-limited attacks proposes to attack facial recognition neural nets with the adversarial eyeglasses \cite{AdvA-RealFaceID-Sharif1}. The authors propose a method to print adversarial perturbation on the eyeglasses frame with the help of Total Variation (TV) loss and non-printability score (NPS). TV loss is designed to make the image more smooth. Thus it makes an attack more stable for different image interpolation methods on the devices and makes it more inconspicuousness for human. NPS is designed to deal with the difference in digital RGB-values and the ability of real printers to reproduce these values.

In general, most of the subsequent works for the real-world attack use the concepts of $L_{0}$-limited perturbation, EOT, TV loss, and NPS. Let us briefly list them. In \cite{AdvA-RealClass-Sign-Eykholt} the authors construct the physical attack for the traffic sign recognition model using EOT and NPS for making either adversarial posters (attacking the whole traffic sign area) or adversarial stickers (black and white stickers on the real traffic sign). The works of \cite{AdvA-RealClass-Sign-Sitawarin1, AdvA-RealClass-Sign-Sitawarin2} use some form of EOT to attack traffic sign recognition model too.

A number of works are devoted to adversarial attacks on traffic sign detectors in the real world. One of the first works \cite{AdvA-RealDet-Sign-FasterRCNN-first} proposes an adversarial attack on Faster R-CNN \cite{FasterRCNN} stop sign detector using a sort of EOT (handcrafted estimation of a viewing map). Several works used EOT, NPS, and TV loss to attack Faster R-CNN, YOLOv2 \cite{YOLOv2} based traffic sign recognition models \cite{AdvA-RealDet-Sign-FasterRCNN,AdvA-RealDetNote-Sign-YOLOv2-Eykholt, AdvA-RealDet-Sign-YOLOv2-Eykholt}.

Another interesting approach \cite{AdvA-RealDet-Sign-YOLOv3-FasterRCNN} uses the concept of nested adversarial examples where separate non-overlapping adversarial perturbations are generated for close and far distances. This attack is designed for Faster R-CNN and YOLOv3 \cite{YOLOv3}.

A few works are devoted to more complex approaches. One of such works \cite{AdvA-RealDet-Person-YOLOv2} proposes to use EOT, NPS, and TV loss for fooling YOLOv2-based person detector. Another one \cite{AdvA-RealFaceID-Sharif2} is devoted to fooling the Face ID system using adversarial generative nets (a sort of GANs \cite{GAN}) where the generator produces the eyeglasses frame perturbation.

\section{Proposed Method}
\label{sec:method}

In the real-use scenario of the Face ID system, not every captured person is known. That is why predicted similarity with the top-1 class should exceed some predefined threshold to treat face as recognized.

The goal of our paper is to create a rectangular image which can be stuck on the hat and induce Face ID system to decrease similarity to ground truth class below the decision threshold.

In order to achieve this goal we use an attack pipeline which can be described as follows: 1) We apply a novel off-plane transformation to the rectangular image which imitates the form of the rectangular image after placing it on the hat. 2) We project the obtained image on the high-quality face image with small perturbations in the projection parameters to make our attack more robust. 3) We transform the obtained image to the standard template of ArcFace input. 4) We reduce the sum of two parameters: TV loss of the initial rectangular image and cosine similarity between the embedding for the obtained image and the anchor embedding calculated by ArcFace. A schema of the whole pipeline is illustrated in \textbf{Figure \ref{fig:pipeline}}.

\begin{figure}[t]
	\begin{minipage}[t]{1.\linewidth}
		\center{\includegraphics[width=\columnwidth]{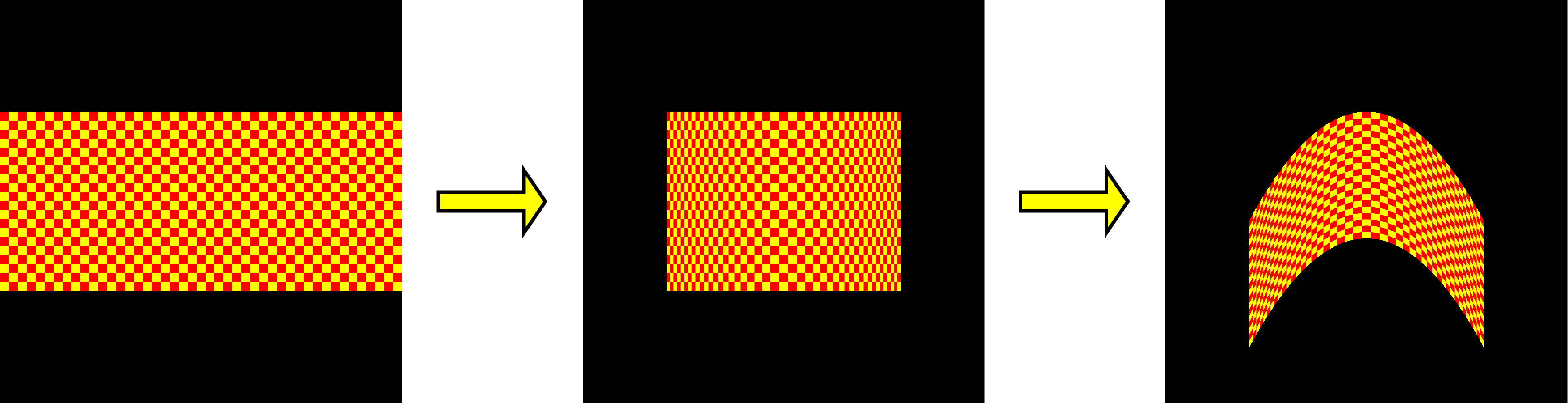}}
	\end{minipage}
	\caption{When we put a rectangular sticker on the hat, it bends and rotates.}
	\label{fig:transformations}
\end{figure}

\subsection{Off-Plain Sticker Transformation}

We split transformations that occur during placing a sticker on the hat into two steps: the off-plane bending of the sticker and pitch rotation of the sticker. Both of these transformations are illustrated in \textbf{Figure \ref{fig:transformations}}.

We simulate the off-plane bending as a parabolic transformation in the 3d space which maps each point of the sticker with initial coordinates $(x,y,0)$ to the new point with coordinates $(x',y,z')$. New points are placed on the parabolic cylinder given by the equation $z=a\cdot x^2$. The origin of all axes is placed in the middle of the sticker. After this transformation, the new $z'$ (off-plane) coordinate of each point of the sticker is equal to $a\cdot x'^2$ and
$$ x' = a \cdot(|x|\cdot\sqrt{x^2+\frac{1}{4\cdot a^2}}+$$
$$+\frac{1}{4\cdot a^2}\cdot\ln{(|x|+\sqrt{x^2+\frac{1}{4\cdot a^2}})}-\frac{1}{4\cdot a^2}\cdot\ln{(\frac{1}{2\cdot a})}).$$
This formula guarantees that the geometric length of the sticker does not change.

To simulate the pitch rotation of the sticker we apply a 3d affine transformation to the obtained coordinates.

We change the parabola rate and the angle of rotation a little during the attack to make the attack more robust since we can not evaluate exact values of these parameters.

\subsection{Sticker Projection}

We use Spatial Transformer Layer (STL) to project the obtained sticker on the image of the face \cite{STN}. We slightly change the parameters of the projection during the attack.

It is crucial to project sticker on the high-quality image of the face. The image interpolation methods, which are applied in the Face ID system and which create a standard template of the face, use the values of neighboring pixels. That is why if we project sticker onto the small face (which is fed to the ArcFace input) then the RGB values on the sticker boundaries differ from that in a real-use scenario since they use values of face pixels too.

\subsection{Final Transformation}

We transform images with the sticker to a standard template for ArcFace using STL. Same as before, we change transformation parameters a little during the attack.

\subsection{Loss Function}

We feed a batch of images obtained by using various parameters to the ArcFace input. The first loss to minimize is a cosine similarity between obtained embeddings $e_{x}$ and some anchor embedding $e_{a}$ for the person:
$$\L_{sim}(x, a) = \cos(e_{x}, e_{a})$$

We minimize a TV loss also concerning the reasons as mentioned earlier. We formulate TV loss as follows:
$${\rm TV}(x)=\sum\limits_{i,j}\left(\left(x_{i,j}-x_{i+1,j}\right)^2+\left(x_{i,j}-x_{i,j+1}\right)^2\right)^\frac{1}{2}$$

The final loss is the weighted sum of the aforementioned losses: 
$$\L_{final}(x, a) = \L_{sim}(x, a) + \lambda\cdot TV(x),$$
where $\lambda$ is a weight for the TV loss ($1e-4$ in our experiments).

We do not use NPS loss since it do not make an influence in our experiments.

\section{Experiments and Results}
\label{sec:experiments}

We use an image of $400\times900$ pixels in our experiments as a sticker image. We project the sticker image to a 600x600 image of the face and then transform it to the 112x112 image.

\subsection{Attack Method}

As stated earlier, we randomly modify images before feeding them to the ArcFace. We construct a batch of generated images and calculate average gradients on the initial sticker using the whole pipeline. We can evaluate gradients in a straight-forward way since each transformation is differentiable.

Note that stickers on each image from the batch are the same during one iteration. Only transformations parameters are different.

We use Iterative FGSM with momentum and several heuristics that were efficient in our experiments.

We split our attack into two stages. During the first stage, we use step value equal to $\frac{5}{255}$ and momentum equal to $0.9$. During the second stage, we use step value equal to $\frac{1}{255}$ and momentum equal to $0.995$. The weight of the TV loss is always equal to $1e-4$.

We use one fixed image with the sticker as a validation where we set up all the parameters to the most real-like looking values. 

We interpolate the last 100 validation values by a linear function using the least square method: after 100 iterations for the first stage and after 200 iterations for the second stage. If the angular coefficient of this linear function is not less than zero then: 1) in the first stage we pass to the second stage of the attack; 2) in the second stage we stop the attack.

\begin{figure}[t]
	\begin{minipage}[h]{0.4\linewidth}
		\center{\includegraphics[width=\columnwidth]{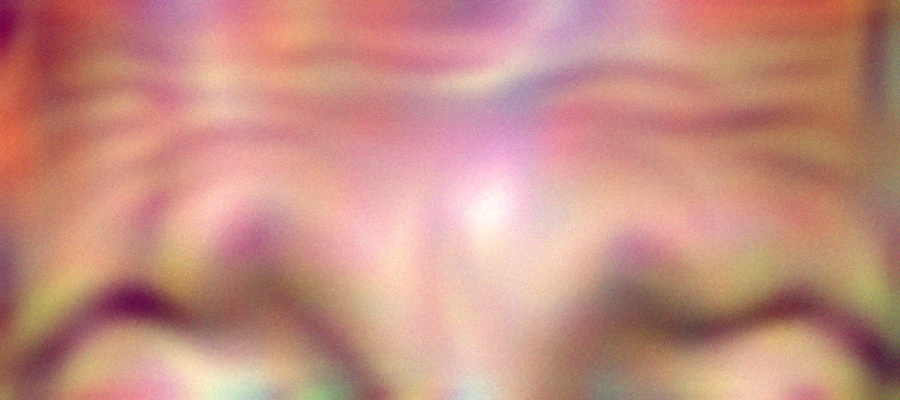}}
	\end{minipage}
	\hfill
	\begin{minipage}[h]{0.4\linewidth}
		\center{\includegraphics[width=\columnwidth]{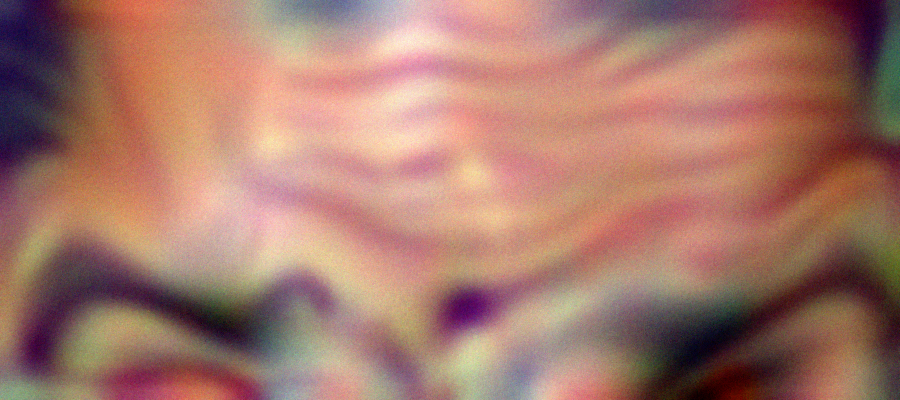}}
	\end{minipage}
	\caption{Examples of the adversarial stickers.}
	\label{fig:stickers}
\end{figure}

\subsection{Sticker Localization}

To find out which place is the best for sticker position, we make two experiments with the sticker localization. First, we attack the image in the digital domain with a sticker placed on various height above the eyez line. It turns out, that the lower placement leads to the better validation values. Further, we change the placement of the sticker after each iteration with respect to the values of gradients on the spatial transformer layer parameters. We limit the possible places for the sticker to make it higher than the eyes. The sticker always moves down to the eyes in our experiments.

Given the above, we put a hat and sticker at the lowest possible position during the attack to achieve the best results. Thus, we put on the hat in our experiments down to the eyes (see \textbf{Figure \ref{fig:example}}).

Some examples of typical adversarial stickers are depicted in \textbf{Figure \ref{fig:stickers}}. It looks like the model draws a raised eyebrows on the sticker. According to the article \cite{Eyebrows}, eyebrows are the most important feature for the face recognition by a human. That is why we believe that some sort of raised eyebrows appears on the sticker. The model draws the most important feature of the face and eyebrows are raised because it is the only reason which makes eyebrows higher than usual.

\begin{figure}[t]
	\begin{minipage}[t]{1.\linewidth}
		\center{\includegraphics[width=\columnwidth]{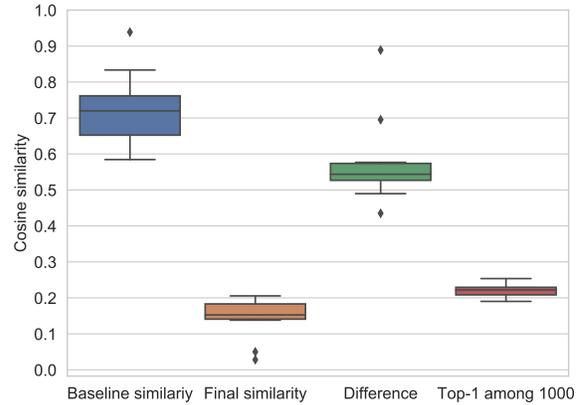}}
	\end{minipage}
	\caption{\textbf{Blue:} Cosine similarities between anchor images and images with a hat. \textbf{Orange:} Cosine similarities between anchor images and images with an adversarial sticker. \textbf{Green:} Differences between aforementioned similarities. \textbf{Red:} The top-1 similarity to the first 1000 classes from CASIA.}
	\label{fig:cosine}
\end{figure}

\subsection{Testing Protocol}

Since it is rather simple to attack a batch of images successfully using the sticker in the digital domain and since the main goal of this paper is to attack the Face ID system in the physical world, we concentrate only on the attacks in the real world. Detailed studies of our approach applied to the digital domain are out of the scope of this work.

At first, we evaluate the success and characteristics of the attacks in the fixed conditions. During this experiment, we use only full-face photos with uniform light. On the next step, we research the robustness of our attack to different angles of the face rotation and light conditions. Finally, we explore the transferability of prepared attacks to other models.

We use first 1000 classes from the CASIA-WebFace dataset as other classes of the recognizer. We do not introduce percent of successful dodging attacks since the success of the attack depend on the threshold which is used in the Face ID system. It can significantly vary according to the purpose of face recognition \footnote{A threshold for ArcFace varies from 0.328 to 0.823 according to the results on IJB-B Still Images Identification test with FAR from $1e-1$ to $1e-3$.}. Instead of the ratio of successes, we explore the following values:
\begin{itemize}
	\item Cosine similarity between ground truth embedding and embedding for a photo with a hat. This is a baseline similarity.
	\item Cosine similarity between ground truth embedding and embedding for a photo with an adversarial sticker. This is a final similarity.
	\item The difference between baseline similarity and final similarity.
	\item The top-1 similarity to the 1000 classes from CASIA.
\end{itemize}

\subsection{Experiments with fixed conditions}

We start with the experiments where all photos and real-world testing are made in the same conditions. We evaluate the values for 10 people with different age and gender: four females of age 30, 23, 16, 5 and six males of age 36, 32, 29, 24, 24, 8. We use 3 photos of each person to create an attack: a simple photo which we need to calculate the ground truth embedding; a photo with the hat which we need to calculate the baseline similarity and obtain the adversarial sticker; a photo with the white sticker on the hat which we need to find parameters of the sticker transformations for this person. Then we print the adversarial sticker for each person and make the fourth photo with this sticker on the hat to obtain final values.

We use boxplot to show the distributions of the obtained values (see \textbf{Figure \ref{fig:cosine}}). As can be seen, adversarial stickers significantly reduce similarity to the ground truth class. Only one attack achieves similarity more than 0.2 in the real world and the same attack achieves similarity to the top-1 class from CASIA less than 0.2. Thus, similarity to the top-1 class from the first 1000 CASIA classes is almost always bigger than similarity to the ground truth although it is not the aim of our attack. It is noteworthy that in most cases the adversarial sticker reduces similarity to the ground truth on more than 0.5. Both attacks that decrease similarity by less than 0.5 relate to children under 10 years old. The baseline similarity is initially smaller for children.

\begin{figure}[t]
	\begin{minipage}[t]{1.\linewidth}
		\center{\includegraphics[width=\columnwidth]{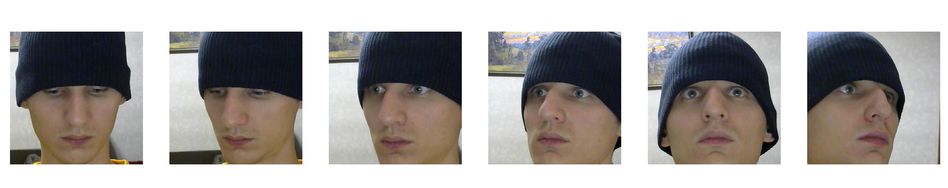}}
	\end{minipage}
	\vfill
	\center\begin{minipage}[t]{0.83\linewidth}
		\center{\includegraphics[width=\columnwidth]{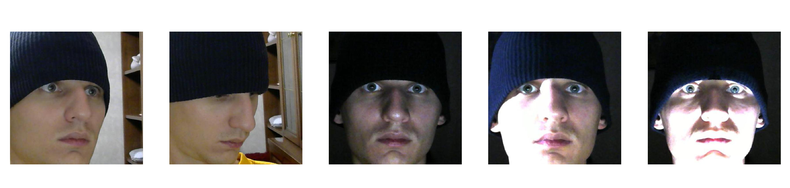}}
	\end{minipage}
	\caption{We make 11 extra photos for some persons to examine the power of the attack in the various conditions. Poses from 1 to 11 are placed from left to right from top to the bottom.}
	\label{fig:conditions}
\end{figure}

\begin{figure}[t]
\begin{minipage}[t]{1.\linewidth}
	\center{\includegraphics[width=\columnwidth]{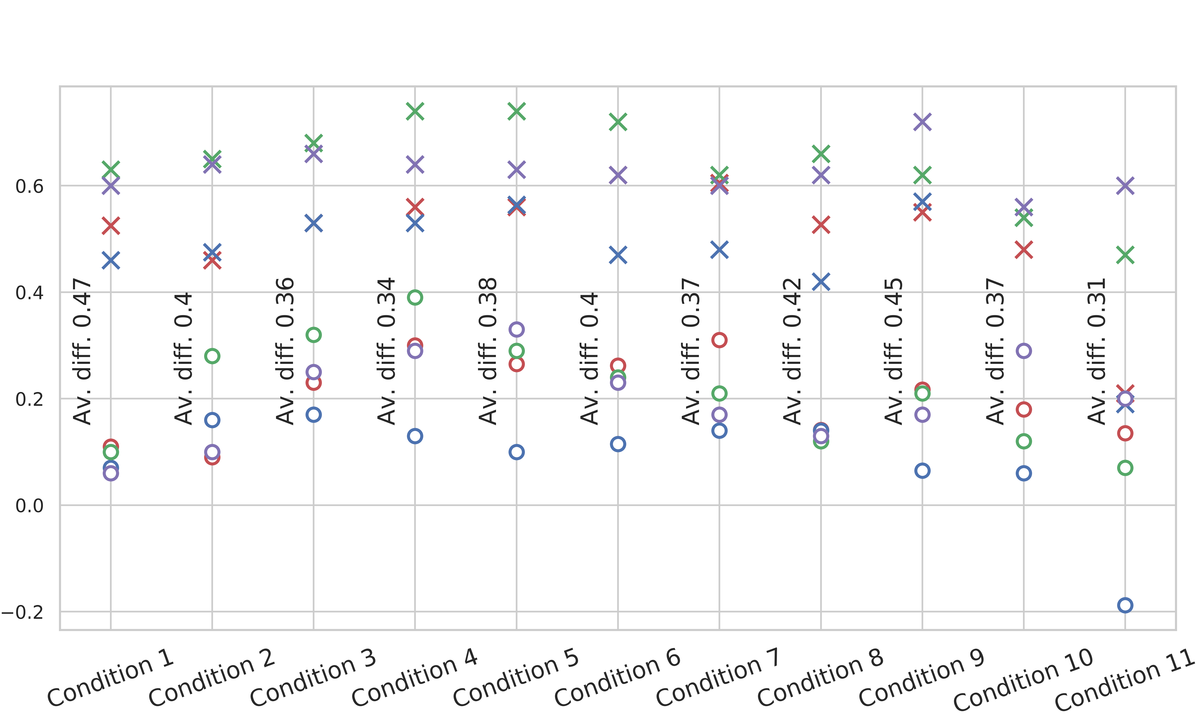}}
\end{minipage}
\caption{Baseline and final similarity for various shooting conditions. Different persons are depicted with different colors. Circle markers are used to show final similarity with adversarial sticker and x markers are used to show baseline similarity.}
\label{fig:various}
\end{figure}

\subsection{Experiments with various conditions}

In order to examine the robustness of our approach to different shooting conditions, we make 22 extra photos for 4 persons from the first 10. These photos consist of 11 pairs. Each pair is made in the same conditions. The first photo of each pair is a photo in a hat that is used to evaluate baseline similarity. The second photo of each pair is a photo in a hat with an adversarial sticker that is used to evaluate the final similarity. 8 pairs correspond to the different combination of head tilts (lean forward, lean back, turn left, turn right) and 3 pairs correspond to different lighting conditions. Examples of shooting conditions are depicted in \textbf{Figure \ref{fig:conditions}}. It is worth noting that we use stickers from the previous step without making new attacks.

The results are illustrated in  \textbf{Figure \ref{fig:various}}. Although final similarity increases, the attack still works. We do not want to jump to a conclusion since the testing set is crucially small but we believe that our approach is robust to rotations of the head that keep sticker visible.

We find out that the bigger area of the sticker on the photo leads to the lower similarity. When the head leans forward, the final similarity is still less than 0.2 and it gradually increases while the head rises. Using better projective and rendering technique and larger adversarial accessories (e.g. using of all area of the hat for the attack) can make you fully unrecognizable for surveillance cameras.

\begin{figure}[t!]
	\begin{minipage}[t]{1.\linewidth}
		\center{\includegraphics[width=\columnwidth]{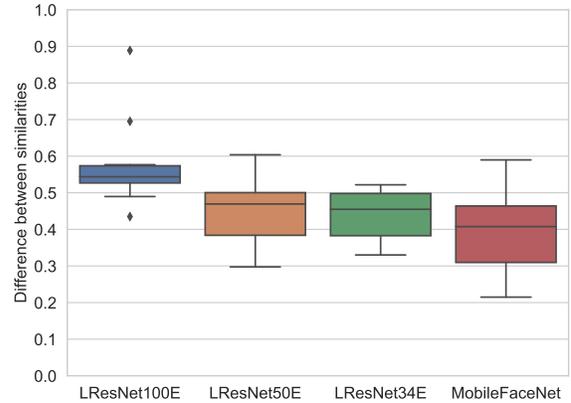}}
	\end{minipage}
	\caption{Differences between baseline and final similarities of one attack on different models. LResNet100E was used to prepare the attack.}
	\label{fig:transferability}
\end{figure}

\subsection{Experiments with transferability}

Finally, we examine the robustness of our attacks to other Face ID models. The models have been taken from InsightFace Model Zoo \cite{ArcFace-model}. These networks have different architectures and they used different loss functions and datasets for training in comparison to the LResNet100E-IR, ArcFace@ms1m-refine-v2.

We use photos from the first experiment to evaluate similarities: a full-face photo, a photo in a hat, a photo with an adversarial sticker on the hat. We calculate the baseline and final similarities for each of 10 persons. The differences between the baseline and final similarities for each model are depicted in \textbf{Figure \ref{fig:transferability}} using boxplots.

We observe that our real-world attack behaves like a usual adversarial attack in the digital domain. Although the strength of the attack decreases, it still makes a person less recognizable.

\section{Conclusion and Future Work}
\label{sec:conclusion}

We have proposed a novel method to attack the Face ID system called AdvHat. Our method can be easily reproducible as well as it can efficiently attack the best public Face ID model in different shooting conditions. Experimental results verified the robustness of our attack to the state-of-the-art Face ID system ArcFace. In the future, we would like to apply our model on state-of-the-art face detectors. 

\section{Acknowledgments}
\label{sec:acknowledgments}

Authors would like to thank their family members and colleagues from Huawei Moscow Research Center for help in conducting experiments, and I. Mazurenko and Y. Xiong from Intelligent Systems Laboratory for guidance and support.


\begin{thebibliography}{10}

\bibitem{Face-Ver-Taigman}
Y.~Taigman, M.~Yang, M.~Ranzato, and L.~Wolf,
\newblock ``Deepface: Closing the gap to human-level performance in face verification'',
\newblock In \textit{Proceedings of the IEEE conference on computer vision and pattern recognition}, pp. 1701--1708 (2014).

\bibitem{Face-ID-Taigman}
Y.~Taigman, M.~Yang, M.~Ranzato, and L.~Wolf,
\newblock ``Web-Scale Training for Face Identification'',
\newblock arXiv preprint arXiv:1406.5266 (2014).

\bibitem{FaceNet}
F.~Schroff, D.~Kalenichenko, and J.~Philbin,
\newblock ``FaceNet: A Unified Embedding for Face Recognition and Clustering'',
\newblock arXiv preprint arXiv:1503.03832 (2015).

\bibitem{CASIA}
D.~Yi, Z.~Lei, S.~Liao, and S.~Li,
\newblock ``Learning Face Representation from Scratch'',
\newblock arXiv preprint arXiv:1411.7923 (2014).

\bibitem{MS-Celeb-1M}
Y.~Guo, L.~Zhang, Y.~Hu, X.~He, and J.~Gao,
\newblock ``MS-Celeb-1M: A Dataset and Benchmark for Large-Scale Face Recognition'',
\newblock arXiv preprint arXiv:1607.08221 (2016).

\bibitem{FaceID-L-Softmax}
W.~Liu, Y.~Wen, Z.~Yu, and M.~Yang,
\newblock ``Large-Margin Softmax Loss for Convolutional Neural Networks'',
\newblock arXiv preprint arXiv:1612.02295 (2016).

\bibitem{FaceID-A-Softmax}
W.~Liu, Y.~Wen, Z.~Yu, M.~Li, B.~Raj, and L.~Song,
\newblock ``SphereFace: Deep Hypersphere Embedding for Face Recognition'',
\newblock arXiv preprint arXiv:1704.08063 (2017).

\bibitem{FaceID-AM-Softmax}
F.~Wang, W.~Liu, H.~Liu, and J.~Cheng,
\newblock ``Additive Margin Softmax for Face Verification'',
\newblock arXiv preprint arXiv:1801.05599 (2018).

\bibitem{FaceID-CosFace}
H.~Wang, Y.~Wang, Z.~Zhou, X.~Ji, D.~Gong, J.~Zhou, Z.~Li, and W.~Liu,
\newblock ``CosFace: Large Margin Cosine Loss for Deep Face Recognition'',
\newblock arXiv preprint arXiv:1801.09414 (2018).

\bibitem{ArcFace-paper}
J.~Deng, J.~Guo, N.~Xue, and S.~Zafeiriou,
\newblock ``Arcface: Additive angular margin loss for deep face recognition'',
\newblock arXiv preprint arXiv:1801.07698 (2018).

\bibitem{Megaface1}
D.~Miller, E.~Brossard, S.~Seitz, and I.~Kemelmacher-Shlizerman,
\newblock ``MegaFace: A Million Faces for Recognition at Scale'',
\newblock arXiv preprint arXiv:1505.02108 (2015).

\bibitem{Megaface2}
I.~Kemelmacher-Shlizerman, S.~Seitz, D.~Miller, and E.~Brossard,
\newblock ``The MegaFace Benchmark: 1 Million Faces for Recognition at Scale'',
\newblock arXiv preprint arXiv:1512.00596 (2015).

\bibitem{NIST-FRVT-1N}
P.~Grother, M.~Ngan, and K.~Hanaoka,
\newblock ``Ongoing Face Recognition Vendor Test (FRVT) Part 2: Identification'',
\newblock NIST Interagency/Internal Report (NISTIR) - 8238.

\bibitem{Deepglint}
Ms-celeb-1m challenge 3: Face feature test/trillion pairs,
\newblock \url{http://trillionpairs.deepglint.com/}.

\bibitem{AdvA-first}
C.~Szegedy, W.~Zaremba, I.~Sutskever, J.~Bruna, D.~Erhan, I.~Goodfellow, and R.~Fergus,
\newblock ``Intriguing properties of neural networks'',
\newblock arXiv preprint arXiv:1312.6199 (2013).

\bibitem{ArcFace-model}
InsightFace Model Zoo,
\newblock LResNet100E-IR, ArcFace@ms1m-refine-v2,
\newblock \url{https://github.com/deepinsight/insightface/wiki/Model-Zoo}.

\bibitem{AdvA-FGSM}
I.~Goodfellow, J. Shlens, and C.~Szegedy,
\newblock ``Explaining and harnessing adversarial examples'',
\newblock arXiv preprint arXiv:1412.6572 (2014).

\bibitem{AdvA-PGD}
A.~Madry, A.~Makelov, L.~Schmidt, D.~Tsipras, and A.~Vladu,
\newblock ``Towards deep learning models resistant to adversarial examples'',
\newblock arXiv preprint arXiv:1706.06083 (2017).

\bibitem{AdvA-Momentum}
Y.~Dong, F.~Liao, T.~Pang, H.~Su, J.~Zhu, X.~Hu, and J.~Li,
\newblock ``Boosting Adversarial Attacks with Momentum'',
\newblock arXiv preprint arXiv:1710.06081 (2017).

\bibitem{AdvA-JSMA}
N.~Papernot, P.~McDaniel, S.~Jha, M.~Fredrikson, Z.~Celik, and A.~Swami,
\newblock ``The limitations of deep learning in adversarial settings'',
\newblock arXiv preprint arXiv:1511.07528 (2015).

\bibitem{AdvA-OnePixel}
J.~Su, D.~Vargas, and S.~Kouichi,
\newblock ``One pixel attack for fooling deep neural networks'',
\newblock arXiv preprint arXiv:1710.08864 (2017).

\bibitem{AdvA-SegmDet}
C.~Xie, J.~Wang, Z.~Zhang, Y.~Zhou, L.~Xie, and A.~Yuille,
\newblock ``Adversarial examples for semantic segmentation and object detection'',
\newblock arXiv preprint arXiv:1703.08603 (2017).

\bibitem{AdvA-BB}
N.~Papernot, P.~McDaniel, I.~Goodfellow, S.~Jha, Z.~Celik, and A.~Swami,
\newblock ``Practical black-box attacks against machine learning'',
\newblock arXiv preprint arXiv:1602.02697 (2016).

\bibitem{AdvA-Universal}
S.~Moosavi-Dezfooli, A.~Fawzi, O.~Fawzi, and P.~Frossard,
\newblock ``Universal adversarial perturbations'',
\newblock arXiv preprint arXiv:1610.08401 (2016).

\bibitem{AdvA-RealClass-Paper-Kurakin}
A.~Kurakin, I.~Goodfellow, and S.~Bengio,
\newblock ``Adversarial examples in the physical world'',
\newblock arXiv preprint arXiv:1607.02533 (2016).


\bibitem{AdvA-RealClass-Transformations-Athalye}
A.~Athalye, and I.~Sutskever,
\newblock ``Synthesizing robust adversarial examples'',
\newblock arXiv preprint arXiv:1707.07397 (2017).

\bibitem{AdvA-AdvPatch}
T.~Brown, D.~Mane, A.~Roy, M.~Abadi, and J.~Gilmer,
\newblock ``Adversarial patch'',
\newblock arXiv preprint arXiv:1712.09665 (2017).

\bibitem{ImageNet}
J.~Deng, W.~Dong, R.~Socher, L.-J.~Li, K.~Li, and L.~Fei-Fei,,
\newblock ``ImageNet: A Large-Scale Hierarchical Image Database'',
\newblock In \textit{Proceedings of the 2009 IEEE Computer Vision and Pattern Recognition}, pp. 248--255 (2009).

\bibitem{AdvA-RealFaceID-Sharif1}
M.~Sharif, S.~Bhagavatula, L.~Bauer, and M.~Reiter,
\newblock ``Accessorize to a crime: Real and stealthy attacks on state-of-the-art face recognition'',
\newblock In \textit{Proceedings of the 2016 ACM SIGSAC Conference on Computer and Communications Security}, pp. 1528--1540 (2016).

\bibitem{AdvA-RealClass-Sign-Eykholt}
K.~Eykholt, I.~Evtimov, E.~Fernandes, B.~Li, A.~Rahmati, C.~Xiao, A.~Prakash, T.~Kohno, and D.~Song,
\newblock ``Robust physical-world attacks on deep learning models'',
\newblock arXiv preprint arXiv:1707.08945 (2017).

\bibitem{AdvA-RealClass-Sign-Sitawarin1}
C.~Sitawarin, A.~Bhagoji, A.~Mosenia, P.~Mittal, and M.~Chiang,
\newblock ``Rogue Signs: Deceiving Traffic Sign Recognition with Malicious Ads and Logos'',
\newblock arXiv preprint arXiv:1801.02780 (2018).

\bibitem{AdvA-RealClass-Sign-Sitawarin2}
C.~Sitawarin, A.~Bhagoji, A.~Mosenia, M.~Chiang, and P.~Mittal,
\newblock ``DARTS: Deceiving Autonomous Cars with Toxic Signs'',
\newblock arXiv preprint arXiv:1802.06430 (2018).

\bibitem{AdvA-RealDet-Sign-FasterRCNN-first}
J.~Lu, H.~Sibai, and E.~Fabry,
\newblock ``Adversarial Examples that Fool Detectors'',
\newblock arXiv preprint arXiv:1712.02494 (2017).

\bibitem{FasterRCNN}
Shaoqing Ren, Kaiming He, Ross Girshick, Jian Sun
\newblock ``Faster R-CNN: Towards Real-Time Object Detection with Region Proposal Networks'',
\newblock arXiv preprint arXiv:1506.01497 (2015).

\bibitem{YOLOv2}
J.~Redmon and A.~Farhadi,
\newblock ``YOLO9000: Better, Faster, Stronger'',
\newblock arXiv preprint arXiv:1612.08242 (2016).


\bibitem{AdvA-RealDet-Sign-FasterRCNN}
S.~Chen, C.~Cornelius, J.~Martin, and D.~Chau,
\newblock ``Robust physical adversarial attack on faster r-cnn object detector'',
\newblock arXiv preprint arXiv:1804.05810 (2018).

\bibitem{AdvA-RealDetNote-Sign-YOLOv2-Eykholt}
K.~Eykholt, I.~Evtimov, E.~Fernandes, B.~Li, D.~Song, T.~Kohno, A.~Rahmati, A.~Prakash, and F.~Tramer,
\newblock ``Note on attacking object detectors with adversarial stickers'',
\newblock arXiv preprint arXiv:1712.08062 (2017).

\bibitem{AdvA-RealDet-Sign-YOLOv2-Eykholt}
K.~Eykholt, I.~Evtimov, E.~Fernandes, B.~Li, A.~Rahmati, F.~Tramer, A.~Prakash, T.~Kohno, and D.~Song,
\newblock ``Physical Adversarial Examples for Object Detectors'',
\newblock arXiv preprint arXiv:1807.07769 (2018).

\bibitem{AdvA-RealDet-Sign-YOLOv3-FasterRCNN}
Y.~Zhao, H.~Zhu, R.~Liang, Q.~Shen, S.~Zhang, and K.~Chen,
\newblock ``Seeing isn't Believing: Practical Adversarial Attack Against Object Detectors'',
\newblock  arXiv preprint arXiv:1812.10217 (2018).

\bibitem{YOLOv3}
J.~Redmon and A.~Farhadi,
\newblock ``YOLOv3: An Incremental Improvement'',
\newblock arXiv preprint arXiv:1804.02767 (2018).

\bibitem{AdvA-RealDet-Person-YOLOv2}
S.~Thys, W.~Ranst, and T.~Goedeme,
\newblock ``Fooling automated surveillance cameras: adversarial patches to attack person detection'',
\newblock  arXiv preprint arXiv:1904.08653 (2019).

\bibitem{AdvA-RealFaceID-Sharif2}
M.~Sharif, S.~Bhagavatula, L.~Bauer, and M.~Reiter,
\newblock ``A General Framework for Adversarial Examples with Objectives'',
\newblock arXiv preprint arXiv:1801.00349 (2018).

\bibitem{GAN}
I.~Goodfellow, J.~Pouget-Abadie, M.~Mirza, B.~Xu, D.~Warde-Farley, S.~Ozair, A.~Courville, and Y.~Bengio,
\newblock ``Generative Adversarial Networks'',
\newblock arXiv preprint arXiv:1406.2661 (2014).

\bibitem{STN}
M.~Jaderberg, K.~Simonyan, and A.~Zisserman,
\newblock ``Spatial transformer networks'',
\newblock In \textit{Advances in neural information processing systems}, pp. 2017--2025 (2015).

\bibitem{Eyebrows}
P.~Sinha, B.~Balas, Y.~Ostrovsky, and R.~Russell,
\newblock ``Face recognition by humans: Nineteen results all computer vision researchers should know about'',
\newblock In \textit{Proceedings of the IEEE}, vol. 94, No 11, pp. 1948--1962 (2006).


\end{thebibliography}
\end{document}